%% file: main.tex
\documentclass[final,12pt]{colt2022} 

\usepackage[utf8]{inputenc} 
\usepackage[T1]{fontenc}    

\newif\ifneurips
\neuripsfalse
\input{macros}
\renewcommand{\cite}{\citep}

\title[SAAC: Safe Reinforcement Learning as an Adversarial Game of Actor-Critics]{SAAC: Safe Reinforcement Learning as an\\ Adversarial Game of Actor-Critics}

\coltauthor{
 \Name{Yannis Flet-Berliac}\thanks{This work was done during Yannis' PhD at Inria Lille (Scool team).} \Email{yfletberliac@cs.stanford.edu}\\
 \addr Computer Science, Stanford University, Stanford, CA, USA
 \AND
 \Name{Debabrota Basu} \Email{debabrota.basu@inria.fr}\\
 \addr Univ. Lille, Inria, CNRS, Centrale Lille, UMR 9189 CRIStAL, F-59000 Lille, France
}

\begin{document}

\maketitle
\input{sections/abstract}
\input{sections/introduction}
\input{sections/background}
\input{sections/saac}
\input{sections/experiments}
\input{sections/conclusion}

\bibliography{references}

\end{document}

%% file: macros.tex
\usepackage{hyperref}       
\usepackage{url}            
\usepackage{booktabs}       
\usepackage{amsfonts,amsmath,amssymb}       
\usepackage{nicefrac}       
\usepackage{microtype}      
\usepackage{xcolor}         
\usepackage{enumitem}
\usepackage{dsfont}
\usepackage{mathtools,algorithm,float}
\usepackage{algorithmic}
\usepackage{wrapfig}
\ifneurips
\usepackage{amsthm}
\fi
\usepackage{tikz}
\usepackage{xspace}
\usetikzlibrary{decorations.pathreplacing,calc}
\newcommand{\tikzmark}[1]{\tikz[overlay,remember picture] \node (#1) {};}

\newcommand*{\AddNote}[4]{%
    \begin{tikzpicture}[overlay, remember picture]
        \draw [decoration={brace,amplitude=0.5em},decorate,thick,blue]
            ($(#3)!(#1.north)!($(#3)-(0,1)$)$) --  
            ($(#3)!(#2.south)!($(#3)-(0,1)$)$)
                node [align=center, text width=2.cm, pos=0.5, anchor=west] {#4};
    \end{tikzpicture}
}%
\newcommand*{\AddNotee}[4]{%
    \begin{tikzpicture}[overlay, remember picture]
        \draw [decoration={brace,amplitude=0.5em},decorate,thick,black]
            ($(#3)!(#1.north)!($(#3)-(0,1)$)$) --  
            ($(#3)!(#2.south)!($(#3)-(0,1)$)$)
                node [align=center, text width=2.cm, pos=0.5, anchor=west] {#4};
    \end{tikzpicture}
}%

\newcommand \mdp {\ensuremath{\mathcal{M}}}
\newcommand \pol {\ensuremath{\pi}}
\newcommand \states {\ensuremath{\mathcal{S}}}
\newcommand \actions {\ensuremath{\mathcal{A}}}
\newcommand \transitions {\ensuremath{\mathcal{T}}}
\newcommand \rewards {\ensuremath{\mathcal{R}}}
\newcommand \agent {\ensuremath{\theta}}
\newcommand \adversary {\ensuremath{\omega}}

\newcommand \real {\ensuremath{\mathbb{R}}}
\newcommand{\algo}{$\mathtt{SAAC}$\xspace}
\newcommand{\algocons}{$\mathtt{SAAC}$-$\mathtt{Cons}$\xspace}
\newcommand{\algocvar}{$\mathtt{SAAC}$-$\mathtt{CVaR}$\xspace}
\newcommand{\algomsd}{$\mathtt{SAAC}$-$\mathtt{MSD}$\xspace}

\newcommand \lag {\ensuremath{\mathcal{L}}}

\newcommand\argmax{\mathop{\rm arg\,max}}
\newcommand\argmin{\mathop{\rm arg\,min}}

\newcommand \expect {\mathop{\mbox{\ensuremath{\mathbb{E}}}}\nolimits}
\newcommand \var {\mathop{\mbox{\ensuremath{\mathbb{V}}}}\nolimits}
\newcommand \KL[2] {D_{\mathrm{KL}}\left( #1 ~\middle\|~ #2\right)}
\newcommand \ent {\mathop{\mbox{\ensuremath{\mathcal{H}}}}\nolimits}
\newcommand \cvar {\mathop{\mbox{\ensuremath{\mathrm{CVaR}}}}\nolimits}
\newcommand \prob[1] {\mathbb{P}\left[#1\right]}
\newcommand \returns {\ensuremath{Z_{\pi}^T(s)}}

\allowdisplaybreaks

%% file: sections/abstract.tex
\begin{abstract}
Although Reinforcement Learning (RL) is effective for sequential decision-making problems under uncertainty, it still fails to thrive in real-world systems where \textit{risk} or \textit{safety} is a binding constraint.
In this paper, we formulate the RL problem with safety constraints as a non-zero-sum game.
While deployed with maximum entropy RL, this formulation leads to a safe adversarially guided soft actor-critic framework, called \algo. In \algo, the adversary aims to break the safety constraint while the RL agent aims to maximize the constrained value function given the adversary's policy. 
The safety constraint on the agent's value function manifests only as a repulsion term between the agent's and the adversary's policies.
Unlike previous approaches, \algo can address different safety criteria such as safe exploration, mean-variance risk sensitivity, and CVaR-like coherent risk sensitivity.
We illustrate the design of the adversary for these constraints.
Then, in each of these variations, we show the agent differentiates itself from the adversary's unsafe actions in addition to learning to solve the task. Finally, for challenging continuous control tasks, we demonstrate that \algo achieves faster convergence, better efficiency, and fewer failures to satisfy the safety constraints than risk-averse distributional RL and risk-neutral soft actor-critic algorithms.
\end{abstract}

%% file: sections/introduction.tex
\section{Introduction}
Reinforcement Learning (RL) is a paradigm of Machine Learning (ML) that addresses the problem of sequential decision making and learning under incomplete information~\cite{puterman2014markov,sutton2018reinforcement}.
Designing an RL algorithm requires both efficient quantification of uncertainty regarding the incomplete information and the probabilistic decision making policy, and effective design of a policy that can leverage these quantifications to achieve optimal performance.
Recent success of RL in structured games, like Chess and Go~\cite{mnih2015human,gibney2016google}, and simulated environments, like continuous control using simulators~\cite{lillicrap2015continuous,degrave2019differentiable}, have drawn significant amount of interest.
Still, real-world deployment of RL in industrial processes, unmanned vehicles, robotics etc., does not only require effectiveness in terms of performance but also being sensitive to risks involved in decisions~\cite{pan2017virtual,dulacarnold2020realworldrlempirical,thananjeyan2021recovery}.
This has motivated a surge in works quantifying risks in RL and designing risk-sensitive (or robust, or safe) RL algorithms~\cite{garcia2015comprehensive,pinto2017robust,ray2019benchmarking,wachi2020safe,eriksson2021sentinel,eysenbach2021maximum}.

\noindent\textbf{Risk-sensitive RL.} In risk-sensitive RL, the perception of risk-sensitivity or safety is embedded mainly using two approaches. The first approach is constraining the RL algorithm to converge in a restricted, “safe” region of the state space~\cite{geibel2005risk,thananjeyan2021recovery,koller2018learning,ray2019benchmarking}. Here, the “safe” region is the part of the state space that obeys some external risk-based constraints, such as the non-slippery part of the floor for a walker. RL algorithms developed using this approach either try to construct policies that generate trajectories which stay in this safe region with high probability~\cite{geibel2005risk}, or to start with a conservative “safe” policy and then to incrementally estimate the maximal safe region~\cite{7798979}.

The other approach is to define a risk-measure on the long-term cumulative return of a policy for a fixed environment, and then to minimize the corresponding total risk~\cite{howard1972risk,garcia2015comprehensive,prashanth2018risk}. A risk-measure is a statistics computed on the cumulative return which quantifies either the spread of the return distribution around its mean value or the heaviness of this distribution's tails~\cite{szego2004risk}. Example of such risk measures are variance, conditional value-at-risk (CVaR)~\cite{rockafellar2000optimization}, exponential utility~\cite{howard1972risk}, variance~\cite{prashanth2016variance}, etc. These risk-measures are also extensively used in dynamic pricing~\cite{lim2007relative}, financial decision making~\cite{artzner1999coherent}, robust control~\cite{chen2005risk}, and other decision making problems where risk has consequential effects.

\noindent\textbf{Our Contributions.} In this paper, we unify both of these approaches as a constrained RL problem, and further derive an equivalent non-zero-sum (NZS) stochastic game formulation~\cite{sorin1986asymptotic} of it.
In our NZS game formulation, \textit{risk-sensitive RL reduces to a game between an agent and an adversary} (Sec.~\ref{sec:problem}). The adversary tries to break the \textit{safety constraints}, i.e., either to move out of the “safe” region or to increase the risk measures corresponding to a given policy. In contrast, the agent tries to construct a policy that maximizes its expected long-term return given the adversarial feedback, which is a statistics computed on the adversary's constraint breaking.

Given this formulation, we propose a generic actor-critic framework where any two compatible actor-critic RL algorithms are employed to enact as the agent and the adversary to ensure risk-sensitive performance (Sec.~\ref{sec:method}). In order to instantiate our approach, we propose a specific algorithm, \textit{Safe Adversarially guided Actor-Critic} (\algo), that deploys two Soft Actor-Critics (SAC)~\cite{haarnoja2018soft} as the agent and the adversary. We further derive the policy gradients for the two SACs, showing that the risk-sensitivity of the agent is ensured by a term repulsing it from the adversary in the policy space. Interestingly, this term can also be used to seek risk and explore more.

In Sec.~\ref{sec:experiments}, we experimentally verify the risk-sensitive performance of \algo under safe region, CVaR, and variance constraints for continuous control tasks from real-world RL suite~\cite{dulacarnold2020realworldrlempirical}. We show that \algo is not only risk-sensitive but also outperforms the state-of-the-art risk-sensitive RL and distributional RL algorithms.

%% file: sections/background.tex
\section{Background}
In this section, we elaborate the details of the three main components of our work: Markov Decision Process (MDP), Maximum-Entropy RL, and risk-sensitive RL.
\subsection{Markov Decision Process (MDP)}
We consider the RL problems that can be modelled as a \textit{Markov Decision Process (MDP)}~\cite{sutton2018reinforcement}. An MDP is defined as a tuple $\mdp \triangleq \left( \states, \actions, \rewards, \transitions, \gamma \right)$. $\states \subseteq \real^d$ is the \textit{state space}. $\actions$ is the admissible \textit{action space}. $\rewards: \states \times \actions \rightarrow \real$ is the \textit{reward function} that quantifies the goodness or badness of a state-action pair $(s,a)$. $\transitions: \states \times \actions \rightarrow \Delta_{\states}$ is the \textit{transition kernel} that dictates the probability to go to a next state given the present state and action.
Here, $\gamma \in (0,1]$ is the \textit{discount factor} that affects how much weight is given to future rewards.
The goal of the agent is to compute a \textit{policy} $\pi: \states \rightarrow \Delta_{\actions}$ that maximizes the expected value of cumulative rewards obtained by a time horizon $T \in \mathbb{N}$. For a given policy $\pol$, the \textit{value function} or the expected value of discounted cumulative rewards is
\begin{align*}
    V_{\pol}(s) &\triangleq \underset{\underset{s_t \sim \transitions(s_{t-1},a_{t-1})}{a_t \sim \pol(s_t)}}{\expect}\left[\sum_{t=0}^T \gamma^t \rewards(s_t, a_t)|s_0 = s\right]\triangleq \expect_{\pol\mdp}[\returns].
\end{align*}
We refer to $\returns$ as the \textit{return} of policy $\pi$ up to time $T$ and $Q_{\pol}(s,a)$ as the action-value function which is the expected return starting from state $s$, taking action $a$ and following policy $\pol$.

\subsection{Maximum-Entropy RL}
In this paper, we adopt the Maximum-Entropy RL (MaxEnt RL) framework~\cite{eysenbach2019if,eysenbach2021maximum}, also known as entropy-regularized RL~\cite{neu2017unified}. 
In MaxEnt RL, we aim to maximize the sum of value function and the conditional action entropy, $\ent_{\pol}(a|s)$, for a policy $\pol$:
\begin{align*}
    &\argmax_{\pol} \quad V_{\pol}(s) + \ent_{\pol}(a|s)= \underset{\underset{s_t \sim \transitions(s_{t-1},a_{t-1})}{a_t \sim \pol(s_t)}}{\expect}\left[\returns - \log \pol(a_t|s_t) \mid s_0 = s\right].
\end{align*}
Unlike the classical value function maximizing RL that always has a deterministic policy as a solution~\cite{puterman2014markov}, MaxEnt RL tries to learn stochastic policies such that states with multiple near-optimal actions have higher entropy and states with single optimal action have lower entropy.
Interestingly, solving MaxEnt RL is equivalent to computing a policy $\pol$ that has minimum KL-divergence from a target trajectory distribution $\transitions\circ\rewards$:
\begin{equation}
    \argmax_{\pol} V_{\pol}(s) + \ent_{\pol}(a|s)
    = \argmin_{\pol} \KL{\pol(\tau)}{\transitions\circ\rewards(\tau)}.\label{eq:kl_equiv}
\end{equation}
Here, $\tau$ is a trajectory $\lbrace (s_0, a_0), \ldots, (s_T, a_T)\rbrace$. Target distribution $\transitions\circ\rewards$ is a Boltzmann distribution (or softmax) on the cumulative rewards given the trajectory: $\transitions\circ\rewards(\tau) \propto p_0(s) \prod_{t=0}^T \transitions(s_{t+1}|s_t, a_t) \exp[\returns]$. Policy distribution is the distribution of generating trajectory $\tau$ given the policy $\pol$ and MDP $\mdp$: $\pol(\tau) \propto p_0(s) \prod_{t=0}^T \transitions(s_{t+1}|s_t, a_t) \pi(a_t|s_t)$.
Thus in MaxEnt RL, the optimal policy is a Boltzmann distribution over the expected future return of state-action pairs. 

This perspective of MaxEnt RL allows us to design \algo which transforms the robust RL into an adversarial game in the softmax policy space.
MaxEnt RL is widely used in solving complex RL problems as: it enhances exploration~\cite{haarnoja2018soft}, it transforms the optimal control problem in RL into a probabilistic inference problem~\cite{todorov2007linearly,toussaint2009robot}, and it modifies the optimization problem by smoothing the value function landscape~\cite{williams1991function,ahmed2019understanding}. 

\noindent\textbf{Soft Actor-Critic (SAC)~\cite{haarnoja2018soft}.} Specifically, we use the SAC framework to solve the MaxEnt RL problem. 
Following the actor-critic methodology, SAC uses two components, an actor and a critic, to iteratively maximize $V_{\pol}(s) + \ent_{\pol}(a|s)$.
The critic minimizes the soft Bellman residual with a functional approximation $Q_{\phi}$: 
\begin{align}
    J(Q_\phi)=&\mathbb{E}_{\left(s_{t}, a_{t}\right) \sim \mathcal{D}}\Big[\frac{1}{2}\Big(Q_{\phi}\left(s_{t}, a_{t}\right)
    -\left(\rewards\left(s_{t}, a_{t}\right)+\gamma \mathbb{E}_{s_{t+1} \sim \rho}\left[V_{\bar{\phi}}\left(s_{t+1}\right)\right]\right)\Big)^2 \Big],\label{eq:softq}
\end{align}
where $\rho$ is the state marginal of the policy distribution, and $V_{\bar{\phi}}\left(s_{t}\right)\triangleq \mathbb{E}_{a_{t} \sim \pi_{\theta}}\left[Q_{\bar{\phi}}\left(s_{t}, a_{t}\right)-\alpha \log \pi\left(a_{t} | s_{t}\right)\right]$.
Eq.~\eqref{eq:softq} makes use of a target soft Q-function with parameters $\bar{\phi}$ obtained using an exponentially moving average of the soft Q-function parameters $\phi$.~\cite{mnih2015human} has demonstrated this technique stabilizes training.
Given the $Q_{\phi}$, the actor learns the policy parameters $\theta$ by minimizing $J(\pi_{\theta})$:
\begin{equation}
    J(\pi_{\theta})=\mathbb{E}_{s_{t} \sim \mathcal{D}}\left[\mathbb{E}_{a_{t} \sim \pi_{\theta}}\left[\alpha \log \left(\pi_{\theta}\left(a_{t} | s_{t}\right)\right)-Q_{\phi}\left(s_{t}, a_{t}\right)\right]\right].
    \label{eq:policysac}
\end{equation}
Here, $\alpha$ is called the entropy temperature; it regulates the relative importance of the entropy term versus the reward and produces better results. We use the version of SAC with an automatic temperature tuning scheme for $\alpha$.

\subsection{Safe RL}
\textbf{Risk Measure for Safety.} Safe or risk-sensitive RL with MDPs is first considered in~\cite{howard1972risk}, where they aim to maximize an exponential utility function over the cumulative reward: $V_{\pol}(s|\lambda) = \lambda^{-1} \log \expect[\exp(\lambda \returns)]$. This is equivalent to maximizing $V_{\pol}(s)+\lambda \var[\returns]$, such that the high variance in return is penalized for $\lambda<0$ and encouraged for $\lambda>0$. 
Though this approach of using exponential utility in risk-sensitive discrete MDPs dominates the initial phase of safe RL research~\cite{marcus1997risk,coraluppi1999risk,garcia2015comprehensive}, with the invent of coherent risks~\cite{artzner1999coherent}\footnote{Variance is not a coherent risk but standard deviation is.}, researchers have looked into other risk measures, such as Conditional Value-at-Risk (CVaR)\footnote{$\text{CVaR}_{\lambda}$ quantifies expectation of the lowest $\lambda\%$ of a probability distribution~\cite{rockafellar2000optimization}.}~\cite{chow2015risk}. Also, application of RL to large scale problems~\cite{chow2014algorithms,chow2015riskconstrained}, tried to make the algorithms scalable and to extend to the continuous MDPs~\cite{ray2019benchmarking}. Our approach is flexible to consider all these risk measures and both discrete and continuous MDP settings.

\noindent\textbf{Safe Exploration.} Another approach is to consider a part of the state-space to be “safe” and constrain the RL algorithm to explore inside it with high probability. \cite{geibel2005risk} considered a subset of terminal states as “error” states $\mathcal{E} \subseteq \states$ and developed a constrained MDP problem to avoid reaching it:
\begin{align}
    \argmax_{\pi} V_{\pol}(s) \text{ s.t. } \forall s \in \states\setminus\mathcal{E}, \rho_{\pol}(s) \leq \delta.\label{eq:safe_exp}
\end{align}
Here, $\rho_{\pol}(s)$ is the total number of times the agent goes to the terminal error states $\mathcal{E}$.
Due to existence of these error states, even a policy with low variance can produce large risks (e.g. falls or accidents)~\cite{ray2019benchmarking}.

The other approach is to use the Lyapunov theory of stability on the value function. This approach computes a compatible Lyapunov function ensuring safety, and then computes a corresponding region of attraction, i.e., a safe region. Given this structure, the goal becomes to compute a safe policy that stays in this safe region with high probability while maximizing the corresponding value function. Given a Lyapunov function and thus, a region of attraction, this approach can also be formulated as Eq.~\eqref{eq:safe_exp} but with a different $\rho$.
In the following section, we express the aforementioned two approaches to safe RL as a constrained MDP.

\paragraph{Robustness with Chance Constraints.} Another family of approaches are developed from the minimax analysis of robustness. In the minimax approach, an agent tries to maximize the value function for the MDP that yields minimum return. Since this approach is worst-case, it is often too conservative in practice and harder to optimize for a plausible family of MDPs in which the MDP of interest is in.
Thus, for a given unknown MDP, a stochastic version~\cite{heger1994consideration} of this problem is developed using chance constraints. In the chance constraint formulation, the agent maximizes the value given that the return is lower than a threshold $\lambda \in \real$ with probability less than or equal to $\delta \in (0,1]$:
\begin{align*}
 \argmax_{\pi} V_{\pol}(s) &\text{ s.t. } \prob{\returns \leq \lambda} \leq \delta.
\end{align*}

As mentioned in~\cite{prashanth2018risk} and~\cite{chow2014algorithms}, safety constraints can be adopted to develop constrained MDP~\cite{altman1999constrained} formulation of risk-sensitive RL. This motivates the constrained MDP formulation.

%% file: sections/saac.tex
\section{Problem Formulation: Safe RL as a Non-Zero Sum Game}\label{sec:problem}
\textbf{Safe RL as Constrained MDP (CMDP).} All of the aforementioned methods to safe RL can be expressed as a CMDP problem that aims to maximize the value function $V_{\pol}$ of a policy $\pol$ while constraining the total risk $\rho_{\pol}$ below a certain threshold $\delta$:
\begin{align}\label{eq:cmdp}
    &\argmax_{\pi} V_{\pol}(s) 
    \text{ s.t. } \rho_{\pol}(s) \leq \delta \text{ for } \delta >0.
\end{align}
\begin{itemize}[leftmargin=*]
    \item If Mean-Standard Deviation (MSD)~\cite{prashanth2016variance} is the risk measure, $\rho_{\pol}(s) \triangleq \expect\left[\returns|\pol, s_0=s\right] + \lambda \sqrt{\var\left[\returns|\pol, s_0=s\right]}$ ($\lambda < 0$).
    \item If CVaR is the risk measure, $\rho_{\pol}(s) \triangleq \cvar_{\lambda}\left[\returns|\pol, s_0=s\right]$ for $\lambda \in [0,1)$.
    \item For the constraint of staying in the “non-error” states $\states\setminus\mathcal{E}$, $\rho_{\pol}(s) \triangleq \expect\left[\sum_{t=0}^T \mathds{1}(s_{t+1} \in \mathcal{E}) |\pol, s_0=s\in \states\setminus\mathcal{E}\right] = \sum_{t=0}^T\mathbb{P}_{\pol}[s_{t+1} \in \mathcal{E}]$ such that $s_0=s$ is a non-error state. We refer to this as \textit{subspace risk} $\mathrm{Risk}(A, \states)$ for $A \subseteq \states$.
\end{itemize}
\textbf{CMDP as a Non-Zero Sum (NZS) Game.} The most common technique to address the constraint optimization in Eq.~\eqref{eq:cmdp} is formulating its Lagrangian:
\begin{equation}\label{eq:lag}
    \lag(\pol, \beta) \triangleq V_{\pol}(s) - \beta_0 \rho_{\pol}(s), \text{ for } \beta_0 \geq 0. 
\end{equation}
For $\beta_0=0$, this reduces to its risk-neutral counterpart. Instead, as $\beta_0\rightarrow\infty$, this reduces to the unconstrained risk-sensitive approach. Thus, the choice of $\beta_0$ is important. We automatically tune it as described in Sec.~\ref{sec:temp}.

Now, the important question is to estimate the risk function $\rho_{\pol}(s)$. Researchers have either solved an explicit optimization problem to estimate the parameter or subspace corresponding to the risk measure, or used a stochastic estimator of the risk gradients. These approaches are poorly scalable and lead to high variance estimates as there is no provably convergent CVaR estimator in RL settings. In order to circumvent these issues, we deploy \textit{an adversary} that aims to maximize the cumulative risk $\rho_{\pol}(s)$ given the same initial state $s$ and trajectory $\tau$ as \textit{the agent} maximizing Eq.~\eqref{eq:lag} and use it as a proxy for the risk constraint in Eq.~\eqref{eq:lag}:
\begin{align}
    &\theta^* \triangleq \argmax_{\theta} \lag(\theta, \beta) = V_{\pol_{\theta}}(s) - \beta_0 V_{\pol_{\omega}}(s),\notag\\
    &\omega^* \triangleq \argmax_{\omega} V_{\pol_{\omega}}(s).\label{eq:nzs}
\end{align}
Here, we consider that the policies of the agent and the adversary are parameterized by $\theta$ and $\omega$ respectively. The value function of the adversary $V_{\pol_{\omega}}(s,\cdot)$ is designed to estimate the corresponding risk $\rho_{\pol}(s)$.
This is a non-zero sum game (NZS) as the objectives of the adversary and the agent are not the same and do not sum up to $0$.
Following this formulation, any safe RL problem expressed as a CMDP (Eq.~\eqref{eq:cmdp}), can be reduced to a corresponding agent-adversary non-zero sum game (Eq.~\eqref{eq:nzs}). The adversary tries to maximize the risk, and thus to shrink the feasibility region of the agent's value function. The agent tries to maximize the regularized Lagrangian objective in the shrinked feasibility region. We refer to this duelling game as \textit{Risk-sensitive Non-zero Sum (RNS)} game.

Given this RNS formulation of Safe RL problems, we derive a MaxEnt RL equivalent of it in the next section. This formulation naturally leads to a dueling soft actor-critic algorithm (\algo) for performing safe RL tasks.

\section{SAAC: Safe Adversarial Soft Actor-Critics}\label{sec:method}
In this section, we first derive a MaxEnt RL formulation of the Risk-sensitive Non-zero Sum (RNS) game. We show that this naturally leads to a duel between the adversary and the agent in the policy space. Following that, we elaborate the generic architecture of \algo, and the details of designing the risk-seeking adversary for different risk constraints. We conclude the section with a note on automatic adjustment of regularization parameters.

\subsection{Risk-sensitive Non-zero Sum (RNS) Game with MaxEnt RL}
In order to perform the RNS game with MaxEnt RL, we substitute the Q-values in Eq.~\eqref{eq:nzs} with corresponding soft Q-values.
Thus, the adversary's objective is maximizing:
\begin{equation*}
    \expect_{\pol_{\omega}}[Q_{\omega}(s,\cdot)] + \alpha_0 \ent_{\pol_{\omega}}(\pi_\omega(.|s))
\end{equation*}
for $\pi_{\omega} \in \Pi_{\omega}$, and the agent's objective is maximizing:
\begin{align}
\begin{split}
&\expect_{\pol_{\theta}}[Q_{\theta}(s,\cdot)] + \alpha_0 \ent_{\pol_{\theta}}(\pi_\theta(.|s))-\beta_0 (\expect_{\pi_{\theta}}[Q_{\omega}(s,\cdot)] +\alpha_0 \ent_{\pol_{\omega}}(\pi_\omega(.|s)))
\end{split}\label{eq:agent1}
\end{align}
for $\pi_{\theta} \in \Pi_{\theta}$.\\

Following the equivalent KL-divergence formulation in policy space, the adversary aims to compute:
\begin{equation}\label{eq:adversary}
   \omega^* = \argmin_{\omega} \KL{\pol_\omega(.|s)}{\exp\left(\alpha_0^{-1}Q_{\omega}(s,\cdot)\right)/Z_{\omega}(s)}.
\end{equation}
Similarly, the agent's objective is to compute:
\begin{align}\label{eq:agent}
    {\theta}^*  &=\argmax_{\theta}~~\expect_{\pol_{\theta}}[Q_{\theta}(s,\cdot)] + \alpha_0(1+\beta_0) \ent_{\pol_{\theta}}(\pi_\theta(.|s))\notag\\ 
    &+ \alpha_0 \beta_0 \expect_{\pi_{\theta}}[\ln(\pi_\omega(.|s)) - \ln \exp[\alpha_0^{-1}Q_{\omega}(s,\cdot)]]
    +\alpha_0\beta_0 \KL{\pi_{\theta}(\cdot|s)}{\pi_{\omega}(\cdot|s)}\notag\\
    &= \argmin_{\theta} \KL{\pol_\theta(.|s)}{\exp\left((\alpha_0(1+\beta_0))^{-1}Q_{\theta}(s,\cdot)\right)/Z_{\theta}(s)}\notag\\
    &-\alpha_0\beta_0 \expect_{\pi_{\theta}}[\ln(\pi_\omega(.|s)) - \ln \exp[\alpha_0^{-1}Q_{\omega}(s,\cdot)]]
    - \alpha_0\beta_0 \KL{\pi_{\theta}(\cdot|s)}{\pi_{\omega}(\cdot|s)}\notag\\
    &= \argmin_{\theta} \KL{\pol_\theta(.|s)}{\exp\left(\alpha^{-1}Q_{\theta}(s,\cdot)\right)/Z_{\theta}(s)}
    - \beta \KL{\pi_{\theta}(\cdot|s)}{\pi_{\omega^*}(\cdot|s)}.
\end{align}

Here, $\alpha = \alpha_0(1+\beta_0)$ and $\beta =\alpha_0\beta_0$. 

The last equality holds true as $\pi_{\omega^*}(.|s)= \exp\left(\alpha_0^{-1}Q_{\omega^*}(s,\cdot)\right)/Z_{\omega^*}(s)$ for the adversary's optimal policy $\pol_{\omega^*}$, and since the optimization is over $\theta$, adding $\ln Z_{\omega}(s)$ does not make a change.

Additionally, for $\omega \neq \omega^*$, the relaxed objective $-(\KL{\pol_\theta(.|s)}{\exp\left(\alpha^{-1}Q_{\theta}(s,\cdot)\right)/Z_{\theta}(s)} - \beta \KL{\pi_{\theta}(\cdot|s)}{\pi_{\omega}(\cdot|s)})$ is a strict lower bound of the goal of the agent in Eq.~\eqref{eq:agent1}. Thus, maximizing the reduced objective is similar to maximizing the lower bound on the actual objective. This is a similar trick adopted in general EM algorithms~\cite{em} for maximizing likelihoods. Thus, not only in asymptotics, but at every step optimizing the reduced objective allows to maximize the agent's risk-sensitive soft Q-value.

Following this reduction, we observe that performing the RNS game with MaxEnt RL is equivalent to performing the traditional MaxEnt RL for adversary with a risk-seeking Q-function $Q_{\adversary}$, and a modified MaxEnt RL for the agent that includes the usual soft Q-function and a KL-divergence term repulsing the agent's policy $\pol_{\agent}$ from the adversary's policy $\pol_{\adversary}$. This behaviour of RNS game in policy space allows to propose a duelling soft actor-critic algorithm, namely \algo, to solve risk-sensitive RL problems.

\subsection{The \algo Algorithm}\label{sec:saac-algo}
\begin{figure*}[ht!]
    \centering
    \includegraphics[width=\textwidth,height=8cm]{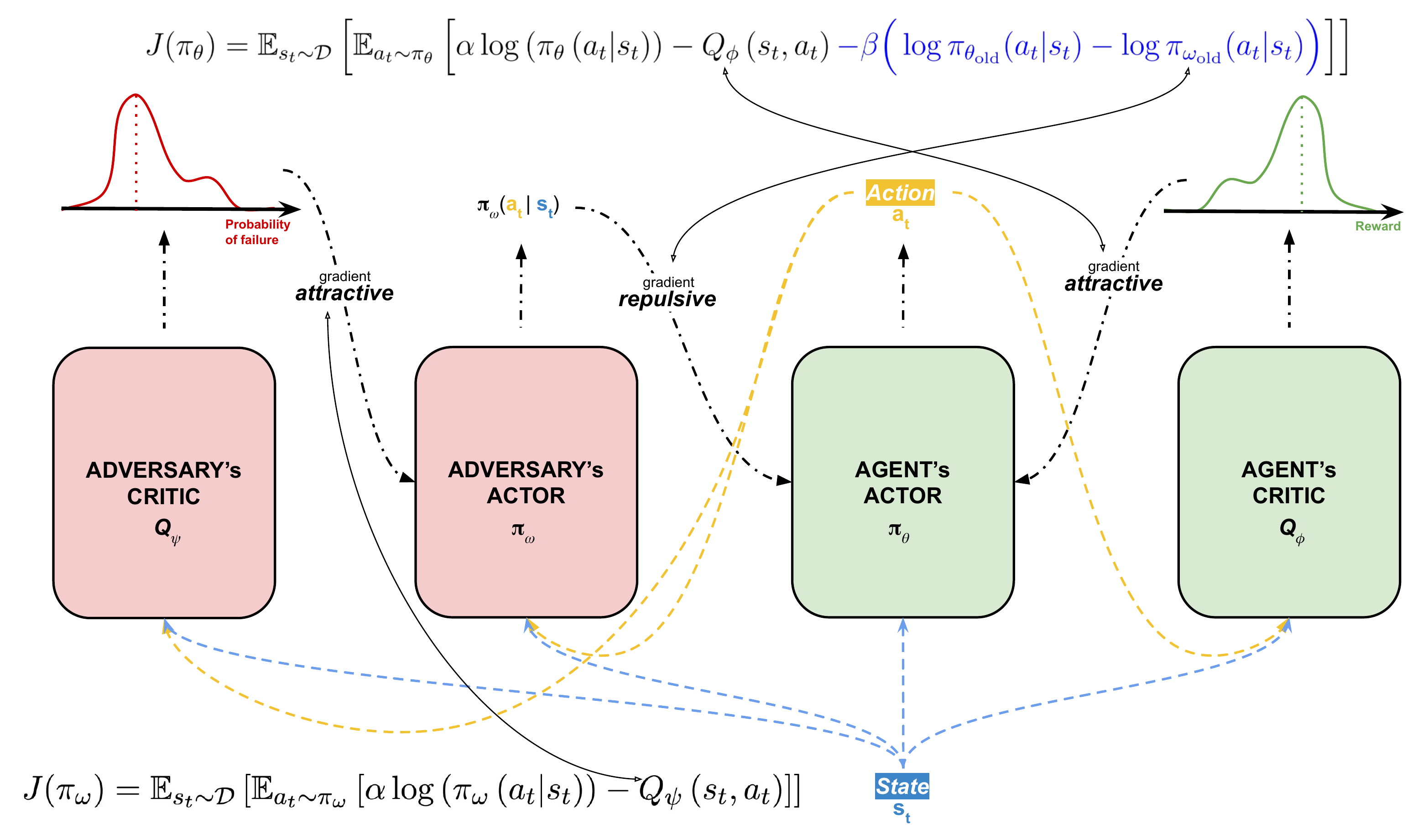}
    \caption{The schematic of the Safe Adversarially guided Actor-Critics (\algo) algorithm.}\label{fig:saac}\vspace*{-1em}
\end{figure*}

We propose an algorithm \algo to solve the objectives of the agent (Eq.~\eqref{eq:agent}) and of the adversary (Eq.~\eqref{eq:adversary}). In \algo, we deploy two soft actor-critics (SACs) to enact the agent and the adversary respectively. We illustrate the schematic of \algo in Fig.~\ref{fig:saac}. 

As a building block for \algo, we deploy the recent version of SAC~\cite{haarnoja2018soft} that uses two soft Q-functions to mitigate positive bias in the policy improvement step in Eq.~\eqref{eq:policysac}, which was encountered in~\cite{hasselt2010double,fujimoto2018addressing}. 
In the design of \algo, we introduce two new ideas: an off-policy deep actor-critic algorithm within the MaxEnt RL framework and a Risk-sensitive Non-zero Sum (RNS) game. \algo engages the agent in safer strategies while finding the optimal actions to \textit{maximize} the expected returns. The role of the adversary is to find a policy that maximizes the probability of breaking the constraints given by the environment. The adversary is trained online with off-policy data given by the agent. We denote the parameter of the adversary policy using $\omega$\footnote{resp. $\omega_\text{old}$ the parameter at the previous iteration.}. For each sequence of transition from the replay buffer, the adversary should find actions that minimize the following loss:
\begin{equation*}
    J(\pi_{\omega})=\mathbb{E}_{s_{t} \sim \mathcal{D}}\left[\mathbb{E}_{a_{t} \sim \pi_{\omega}}\left[\alpha \log \left(\pi_{\omega}\left(a_{t} | s_{t}\right)\right)-Q_{\psi}\left(s_{t}, a_{t}\right)\right]\right].
\end{equation*}

Finally, leveraging the RNS based reduced objective, \algo makes the agent's actor minimize $J(\pi_{\theta})$:
\begin{align*}
    J(\pi_{\theta})=\mathbb{E}_{s_{t} \sim \mathcal{D}}\Big[\mathbb{E}_{a_{t} \sim \pi_{\theta}}\Big[\alpha \log \left(\pi_{\theta}\left(a_{t} | s_{t}\right)\right)-Q_{\phi}\left(s_{t}, a_{t}\right) \textcolor{blue}{- \beta \Big(\log \pi_{\theta_\text{old}}(a_t | s_t) - \log \pi_{\omega_\text{old}}(a_t | s_{t})\Big)}\Big]\Big].
\end{align*}
In \textcolor{blue}{blue} is the repulsion term introduced by \algo. The method alternates between collecting samples from the environment with the current agent's policy and updating the function approximators, namely the adversary's critic $Q_\psi$, the adversary's policy $\pi_\omega$, the agent's critic $Q_\phi$ and the agent's policy $\pi_\theta$. It performs stochastic gradient descent on corresponding loss functions with batches sampled from the replay buffer. We provide a generic description of \algo in Algorithm~\ref{alg:saac}. Now, we provide a few examples of designing the adversary's critic $Q_{\psi}$ for different safety constraints.
\begin{figure}[ht]
  \centering
\vspace{-1em}
\begin{minipage}{\textwidth}
\begin{algorithm}[H]
  \caption{\algo}\label{alg:saac}
\begin{algorithmic}
\STATE \textbf{Input parameters:} $\tau, \lambda_Q, \lambda_\pi, \lambda_\alpha, \lambda_\beta$
\STATE \textbf{Initialize} adversary's and agent's policies and Q-functions parameters $\omega$, $\psi$, $\theta$ and $\phi$
\STATE \textbf{Initialize} temperature parameters $\alpha$ and $\beta$
\STATE $\mathcal{D} \gets \emptyset$
\FOR {each iteration}
\FOR {each step}
\STATE $a_{t} \sim \pi_{\theta}(a_t|s_t)$
\STATE $s_{t+1} \sim {\cal P}\left(s_{t}, a_{t}\right)$
\STATE $\mathcal{D} \gets \mathcal{D} \cup\left\{\left(s_{t}, a_{t}, r_t, s_{t+1}\right)\right\}$
\ENDFOR

\FOR {each gradient step}
\STATE sample batch $\mathcal{B}$ from $\mathcal{D}$
\textcolor{blue}{\STATE$\psi \gets \psi-\lambda_{Q} \hat{\nabla}_{\psi} J_{Q}\left(\psi\right)$\;\;\tikzmark{top}\tikzmark{right}
\STATE$\omega \gets \omega-\lambda_{\pi} \hat{\nabla}_{\omega} J(\pi_{\omega})$}
\textcolor{blue}{\STATE$\beta \gets \beta-\lambda_{\beta} \hat{\nabla}_{\beta} J(\beta)$}
\textcolor{blue}{\STATE$\bar{\psi} \gets \tau \psi+(1-\tau) \bar{\psi}$}\tikzmark{bottom}

\STATE$\phi \gets \phi-\lambda_{Q} \hat{\nabla}_{\phi} J_{Q}\left(\phi\right)$\;\;\,\,\tikzmark{top1}\tikzmark{right1}
\STATE$\theta \gets \theta-\lambda_{\pi} \hat{\nabla}_{\theta} J(\pi_{\theta})$
\STATE$\alpha \gets \alpha-\lambda_{\alpha} \hat{\nabla}_{\alpha} J(\alpha)$
\STATE$\bar{\phi} \gets \tau \phi+(1-\tau) \bar{\phi}$\tikzmark{bottom1}

\ENDFOR
\ENDFOR
\end{algorithmic}
\AddNote{top}{bottom}{right}{Update Adversary}
\AddNotee{top1}{bottom1}{right1}{\;Update Agent}
\end{algorithm}
\end{minipage}
\vspace{-1em}
\end{figure}

\noindent\textbf{\algocons: Subspace Risk.} At every step, the environment signals whether the constraints have been satisfied or not. We construct a reward signal based on this information. This constraint reward, denoted as $r_c$, is $1$ if all the constraints have been broken, and $0$ otherwise. $J(Q_\psi)$ is the soft Bellman residual for the critic responsible with constraint satisfaction:
\begin{align}
\label{eq:softqcritic}
    J(Q_\psi)=\mathbb{E}_{\left(s_{t}, a_{t}\right) \sim \mathcal{D}}\Big[\frac{1}{2}\Big(Q_{\psi}\left(s_{t}, a_{t}\right)-\big(r_c\left(s_{t}, a_{t}\right) +\gamma \mathbb{E}_{s_{t+1} \sim \rho}\mathbb{E}_{a_{t} \sim \pi_{\omega}}\left[Q_{\bar{\psi}}\left(s_{t}, a_{t}\right)-\alpha \log \pi\left(a_{t} | s_{t}\right)\right]\Big)^{2}\Big].
\end{align}

\noindent\textbf{\algomsd: Mean-Standard Deviation (MSD)}. In this case, we consider optimizing a Mean-Standard Deviation risk~\cite{prashanth2016variance}, which we estimate using:
$Q_\psi(s,a) = Q_\phi(s,a) + \lambda\sqrt{\var[Q_\phi(s,a)]}.$
$\lambda<0$ is a hyperparameter that dictates the lower $\lambda-\mathrm{SD}$ considered to represent the lower tail. In the experiments, we use $\lambda=-1$. 
In practice, we approximate the variance $\var[Q_\phi(s,a)]$ using the state-action pairs in the current batch of samples. We refer to the associated method as  \algomsd.

\noindent\textbf{\algocvar: CVaR.} Given a state-action pair $(s,a)$, the Q-value distribution is approximated by a set of quantile values at quantile fractions~\cite{eriksson2021sentinel}. Let $\left\{\tau_{i}\right\}_{i=0, \ldots, N}$ denote a set of quantile fractions, which satisfy $\tau_{0}=0$, $\tau_{N}=1$, $\tau_{i}<\tau_{j}\, \forall i<j$, $\tau_{i} \in[0,1]\, \forall i=0, \ldots, N$, and $\hat{\tau}_{i}=\left(\tau_{i}+\tau_{i+1}\right) / 2$. If $Z^{\pi}: \mathcal{S} \times \mathcal{A} \rightarrow \mathcal{Z}$ denotes the soft action-value of policy $\pi$, $Q_\psi(s,a) = - \sum_{i=0}^{N-1}\left(\tau_{i+1}-\tau_{i}\right) g^{\prime}\left(\hat{\tau}_{i}\right) Z^{\pi_\theta}_{\hat{\tau}_{i}}(s,a;\phi)$
with $g(\tau)=\min \{\tau / \lambda, 1\}$, where $\lambda \in(0,1)$. In the experiments, we set $\lambda=0.25$, i.e. we truncate the right tail of the return distribution by dropping 75\% of the topmost atoms.

\subsection{Automating Adversarial Adjustment}\label{sec:temp}
Similar to the solution introduced in~\cite{haarnoja2018soft}, the adversary temperature $\beta$ and the entropy temperature are automatically adjusted.
Since the adversary bonus can differ across tasks and during training, a fixed coefficient would be a poor solution.
We use $\bar{\mathcal{A}}$ to denote the adversary's bonus target, which is a hyperparameter in \algo. By formulating a constrained optimization problem where the KL-divergence between the agent and the adversary is constrained, $\beta$ is learned by gradient descent with respect to:
\begin{align*}
J(\beta)=\mathbb{E}_{s_{t} \sim \mathcal{D}}\left[\log \beta \cdot\left(\KL{\pi_{\theta}(\cdot|s_t)}{\pi_{\omega}(\cdot|s_t)} -\bar{\mathcal{A}}\right)\right].    
\end{align*}
In addition, the entropy temperature $\alpha$ is also learned by taking a gradient step with respect to the loss:
\[J(\alpha)=\mathbb{E}_{s_{t} \sim \mathcal{D}}\left[\log \alpha \cdot\left(-\log \pi_{\theta}\left(a_{t}|s_{t}\right)-\bar{\mathcal{H}}\right)\right].\]
$\bar{\mathcal{H}}$ is the target entropy: a hyperparameter needed in SAC. We illustrate this in the pseudo-code of SAAC as in Algorithm~\ref{alg:saac}.

%% file: sections/experiments.tex
\section{Experimental Analysis}\label{sec:experiments}
\input{sections/saac_variants}
\textbf{Experimental Setup.}
First, we compare some possible variants of our method. Indeed, as presented in Sec.~\ref{sec:saac-algo}, the adversary has different quantifications of risk to fulfill the objective of finding actions with high probability of breaking the constraints: \algocons, \algocvar, and \algomsd.

Following that, we compare our method with best performing competitors in continuous control problems: SAC~\cite{haarnoja2018soft} and TQC~\cite{kuznetsov2020controlling}. TQC builds on top of C51~\cite{bellemare2017distributional} and QR-DQN~\cite{dabney2018distributional}, and adapt the distributional RL methods for continuous control. Further, they apply truncation for the approximated distributions to control their overestimation and use ensembling on the approximators for additional performance improvement. Finally, we qualitatively compare the behavior of our risk-averse method with that of SAC, using state vectors collected during validation in test environments. Note that for all the experiments (repeated over 9 random seeds), the agents are trained for 1M timesteps and their performance is evaluated at every 1000-th step.

Similar to TQC, we implement \algo on top of SAC and choose to automatically tune the adversary temperature $\beta$ (Sec.~\ref{sec:temp}) and the entropy temperature $\alpha$. Last but not least, using \algo on top of SAC introduces only one hyperparameter: the learning rate for the automatic tuning of $\beta$. All the other hyperparameters are the same as for SAC and are available for consultation in~\cite[Appendix D]{haarnoja2018soft}. For TQC, we employ the same hyperparameters as reported in~\cite{kuznetsov2020controlling}.

\noindent\textbf{Description of Environments.}
To validate the framework of a RNS Game with MaxEnt RL, we conduct a set of experiments in the DM control suite~\cite{tassa2018deepmind}. More specifically, we use the real-world RL challenge\footnote{\href{https://github.com/google-research/realworldrl_suite}{https://github.com/google-research/realworldrl\_suite}}~\cite{dulacarnold2020realworldrlempirical}, which introduces a set of real-world inspired challenges. In this paper, we are particularly interested in the tasks, where a set of constraints are imposed on existing control domains. In the following, we give a short description of the tasks and safety constraints used in the experiments, with their respective observation ($\mathcal{S}$) and action ($\mathcal{A}$) dimensions. First, \textit{realworldrl-walker-walk} ($\mathcal{S}\times\mathcal{A} = 18 \times 6$) corresponds to the dm-control suite \textit{walker} task with (a) joint-specific constrains on the joint angles to be within a range and (b) a constrain on the joint velocities to be within a range. Next, \textit{realworldrl-quadruped-joint-walk} ($\mathcal{S}\times\mathcal{A} = 78 \times 12$) corresponds to the dm-control suite \textit{quadruped} task with the same set of constraints as just described. \textit{realworldrl-quadruped-upright-walk} has a constrain on the quadruped's torso's z-axis to be oriented upwards, and \textit{realworldrl-quadruped-force-walk} limits foot contact forces when touching the ground.
\input{sections/tables}\vspace*{-.8em}
\subsection{Comparison between Risk Quantifiers of \algo}
First, we compare the different variants of \algo allowed by the method's framework in the \textit{realworldrl-walker-walk-returns} task. From Table~\ref{tab:comparison} and Fig.~\ref{fig:constraint1} (lines are average performances and shaded areas represent one standard deviation) we evaluate how our method affects the performance and risk aversion of agents.

In addition to the rate at which the maximum average return is reached by each of the methods compared to SAC, we compare the cumulative number of failures of the agents (the lower the better). As expected, risk-sensitive agents such as \algo decrease the probability of breaking safety constraints. Concurrently, they achieve the maximum average return with much higher sample efficiency, \algomsd ahead. Henceforth, we use the \algomsd version of our method to compare with the baselines.

\subsection{Comparison of \algo to Baselines}
Now, we compare the best performing \algo variant \algomsd with SAC~\cite{haarnoja2018soft}, TQC~\cite{kuznetsov2020controlling} and TQC-CVaR, i.e. an extension of TQC with 16\% of the topmost atoms dropped (cf. Table 6 in~\cite[Appendix B]{kuznetsov2020controlling}) of all Q-function atoms. In Table~\ref{tab:quadruped-upright} and Fig.~\ref{fig:constraint2}, we evaluate \algomsd in \textit{realworldrl-quadruped-upright-walk}. In Table~\ref{tab:quadruped-joint} and Fig.~\ref{fig:constraint3}, we report the results for \textit{realworldrl-quadruped-joint-walk}.

Table~\ref{tab:quadruped-joint} shows that \algomsd performs better than all other baselines both in terms of final performance and in terms of finding risk-averse policies. Moreover, although TQC-CVaR exhibits fewer number of failures over the course of learning, it performs slightly worse than its non-truncated counterpart TQC. Table~\ref{tab:quadruped-upright} confirms the advantage of using \algomsd as a risk-averse MaxEnt RL method over the baselines: overall using \algo allows the agents to achieve faster convergence using safer policies during training. Interestingly, TQC achieves the maximum score of the task a bit later than the SAC agent. Nevertheless, TQC-CVaR, its CVaR variant, opens the door for better sample efficiency score with much safer policies.

\subsection{Visualization of Safer State Space Visitation}
\begin{figure*}[t!]
    \centering
    \begin{minipage}{0.48\textwidth}
    	\centering
    	\includegraphics[width=\linewidth]{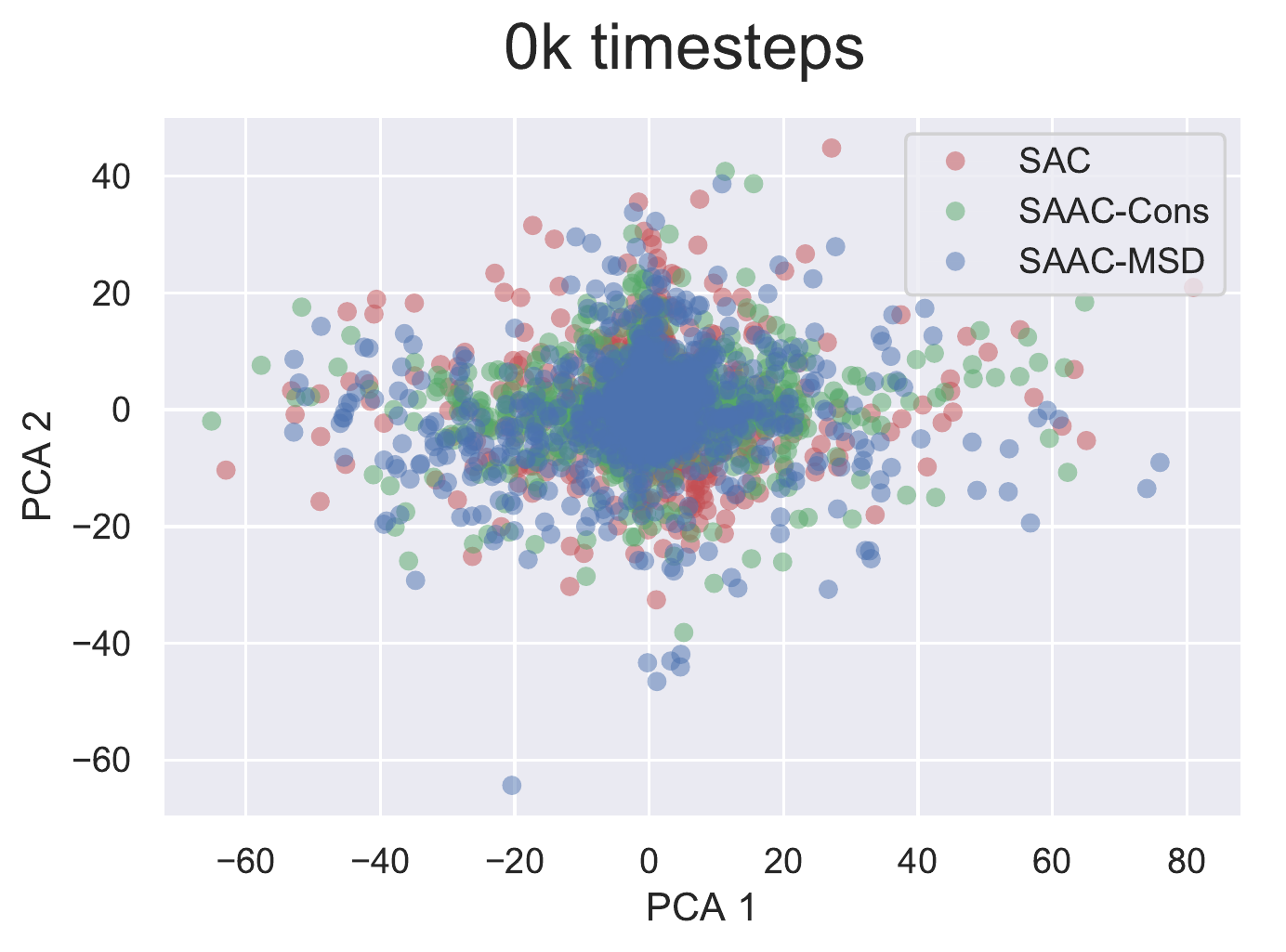}
    \end{minipage}\hfill
\begin{minipage}{0.48\textwidth}
	\centering
	\includegraphics[width=\linewidth]{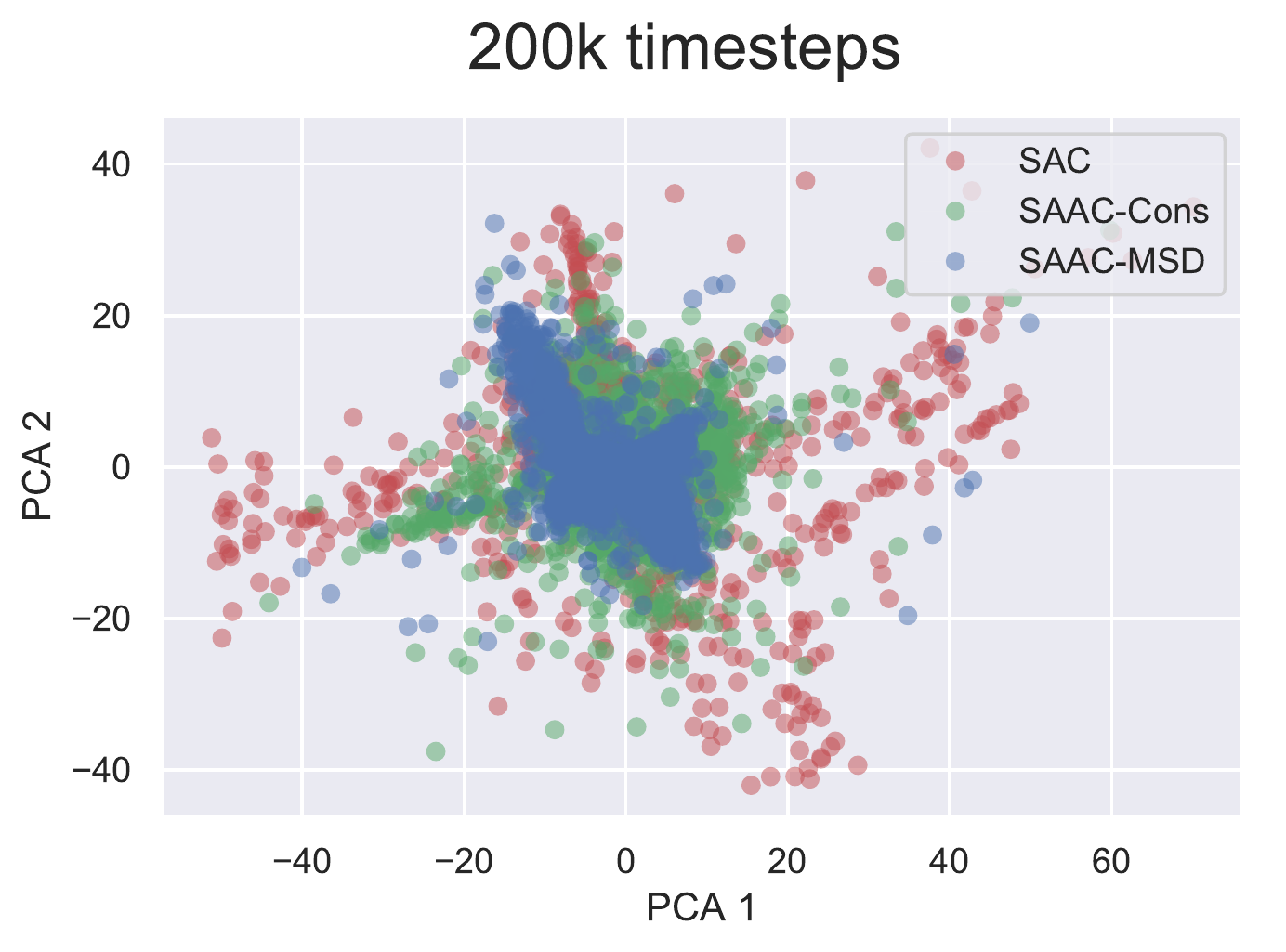}
\end{minipage}\\
  \begin{minipage}{0.48\textwidth}
  	\centering
	\includegraphics[width=\linewidth]{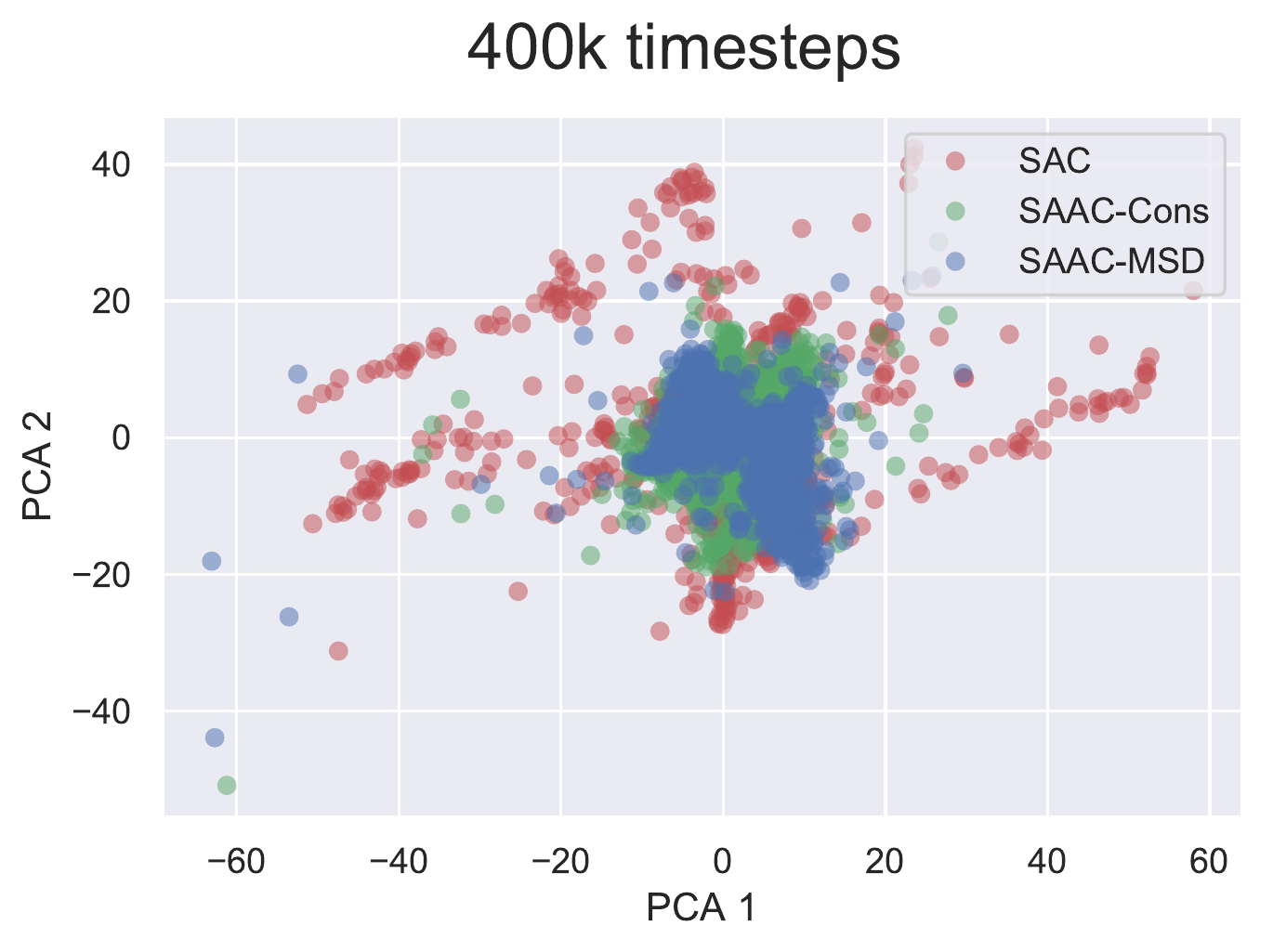}
\end{minipage}\hfill
\begin{minipage}{0.48\textwidth}
	\centering
	\includegraphics[width=\linewidth]{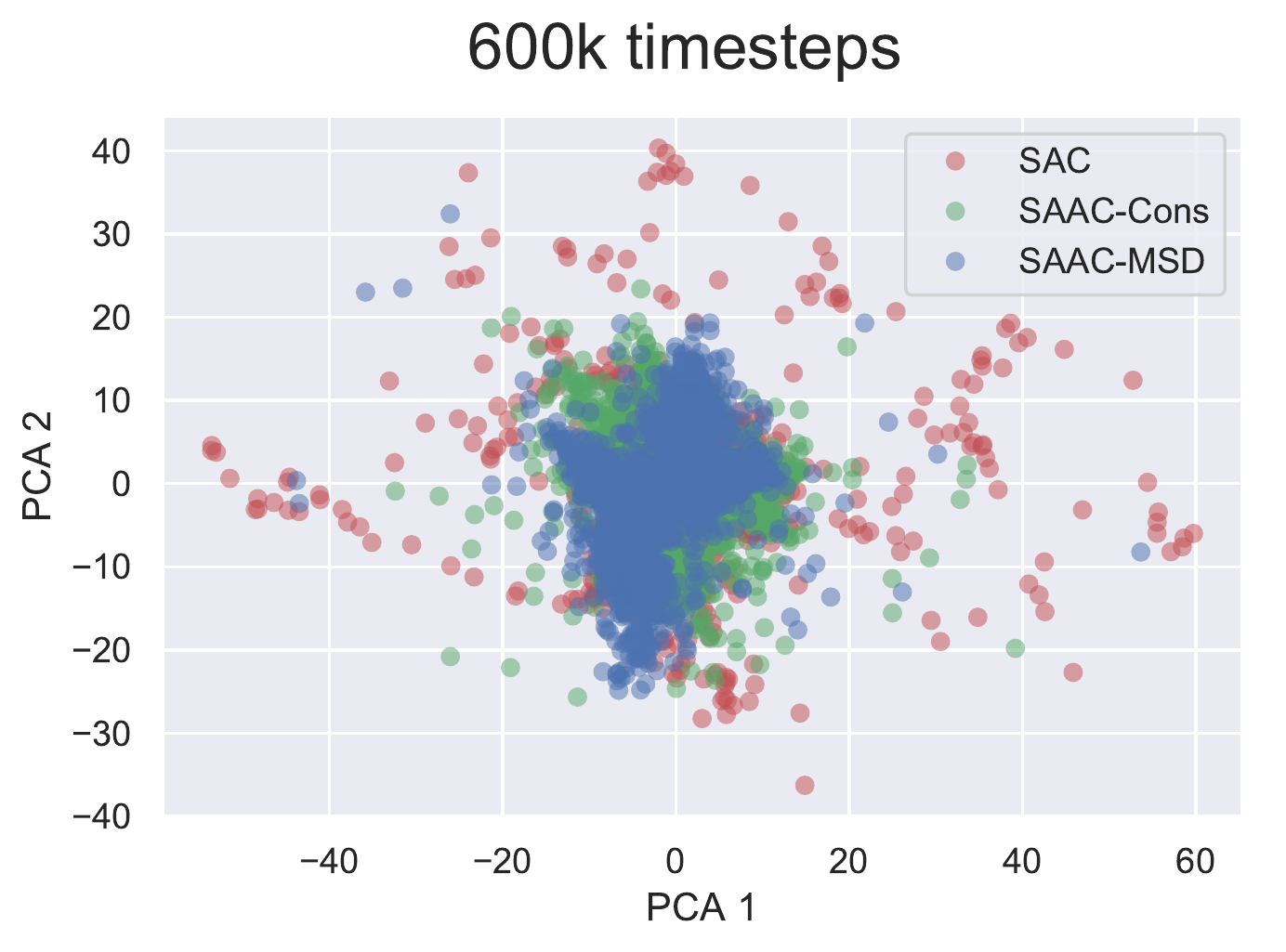}
\end{minipage}
    \caption{Visualization of visited state space projection at different stages of learning in the \textit{realworldrl-walker-walk} task.}\label{fig:state-space}\vspace*{-1em}
\end{figure*}

In this experiment, we choose SAC, \algocons and \algomsd to train a relatively wide spectrum of agents using the same experimental protocol as in Sec. 5.2., and on the \textit{realworldrl-walker-walk} task. We collect samples of states visited during the evaluation phase in a test environment at different stages of the training. The state vectors are projected from a 18D space to a 2D space using PCA. We present the results in Fig.~\ref{fig:state-space}. At the beginning of training, there is no clear distinction in terms of explored state regions, as the learning has not begun yet. On the contrary, during the 200k-600k timesteps, there is a significant difference in terms of state space visitation. In resonance with the cumulative number of failures shown in Fig.~\ref{fig:constraint1}, the results suggest that SAC engages in actions leading to more unsafe states. Conversely, \algo seems to successfully constraint the agents to safe regions.

%% file: sections/saac_variants.tex
\begin{figure*}[t!]
\centering
\begin{minipage}[c]{0.48\linewidth}
	\centering
	{\includegraphics[width=\linewidth]{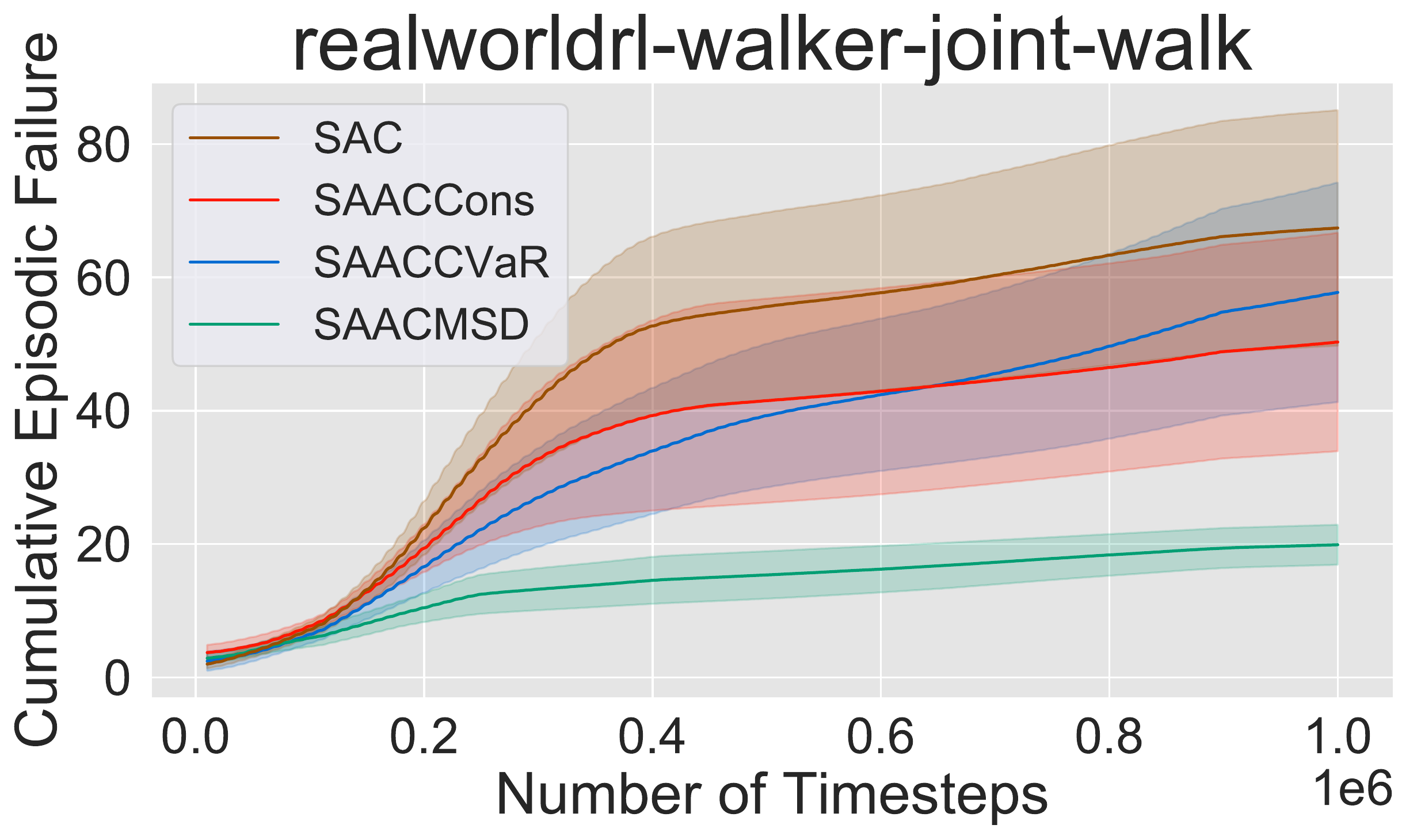}}
    \caption{\algo variants.}
    \label{fig:constraint1}
\end{minipage}\hfill
\begin{minipage}[c]{0.48\linewidth}
	\centering
	{\includegraphics[width=\linewidth]{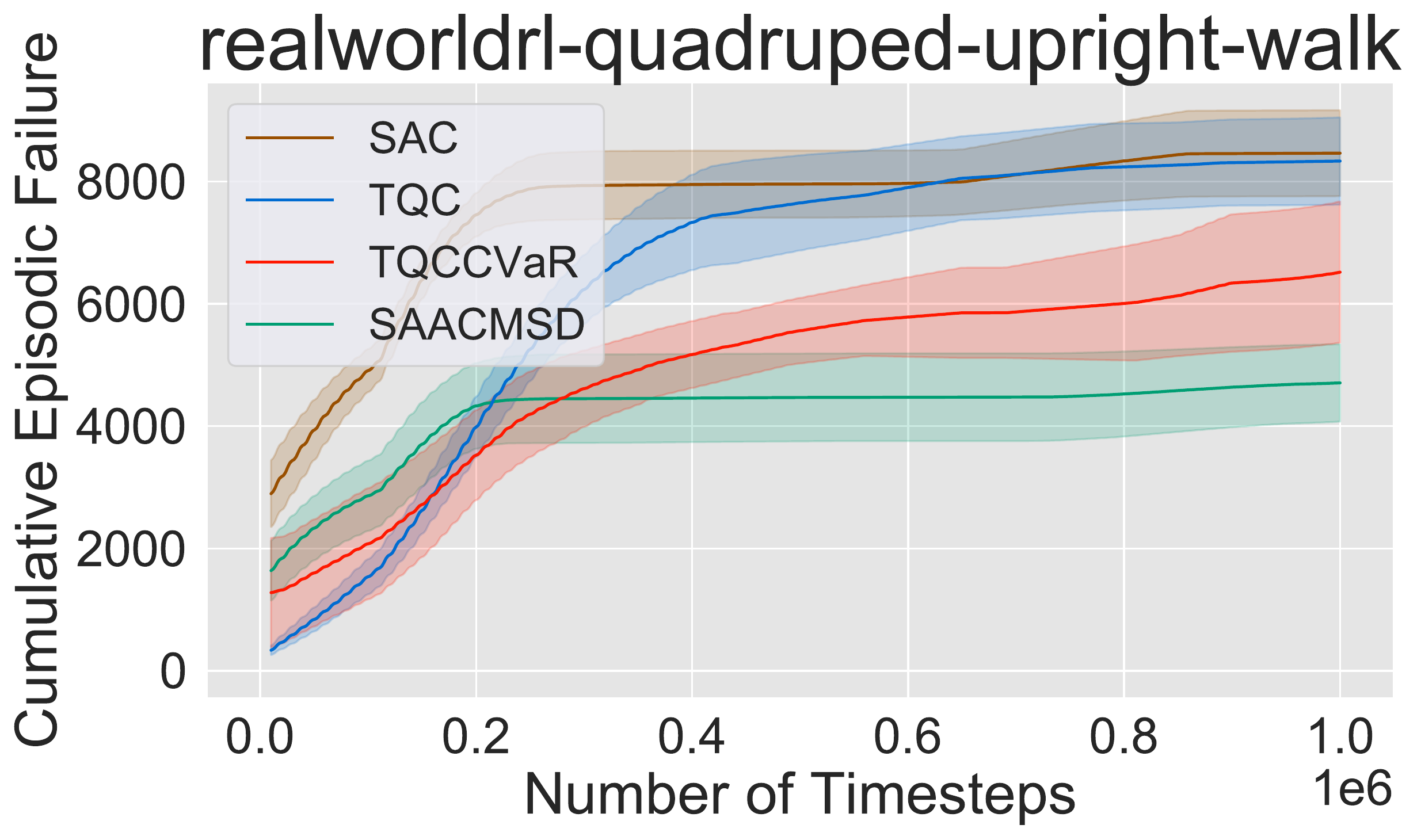}}
    \caption{\small{\algo vs. baselines.}}
    \label{fig:constraint2}
\end{minipage}\\
\begin{minipage}[c]{0.48\linewidth}
	\centering
	{\includegraphics[width=\linewidth]{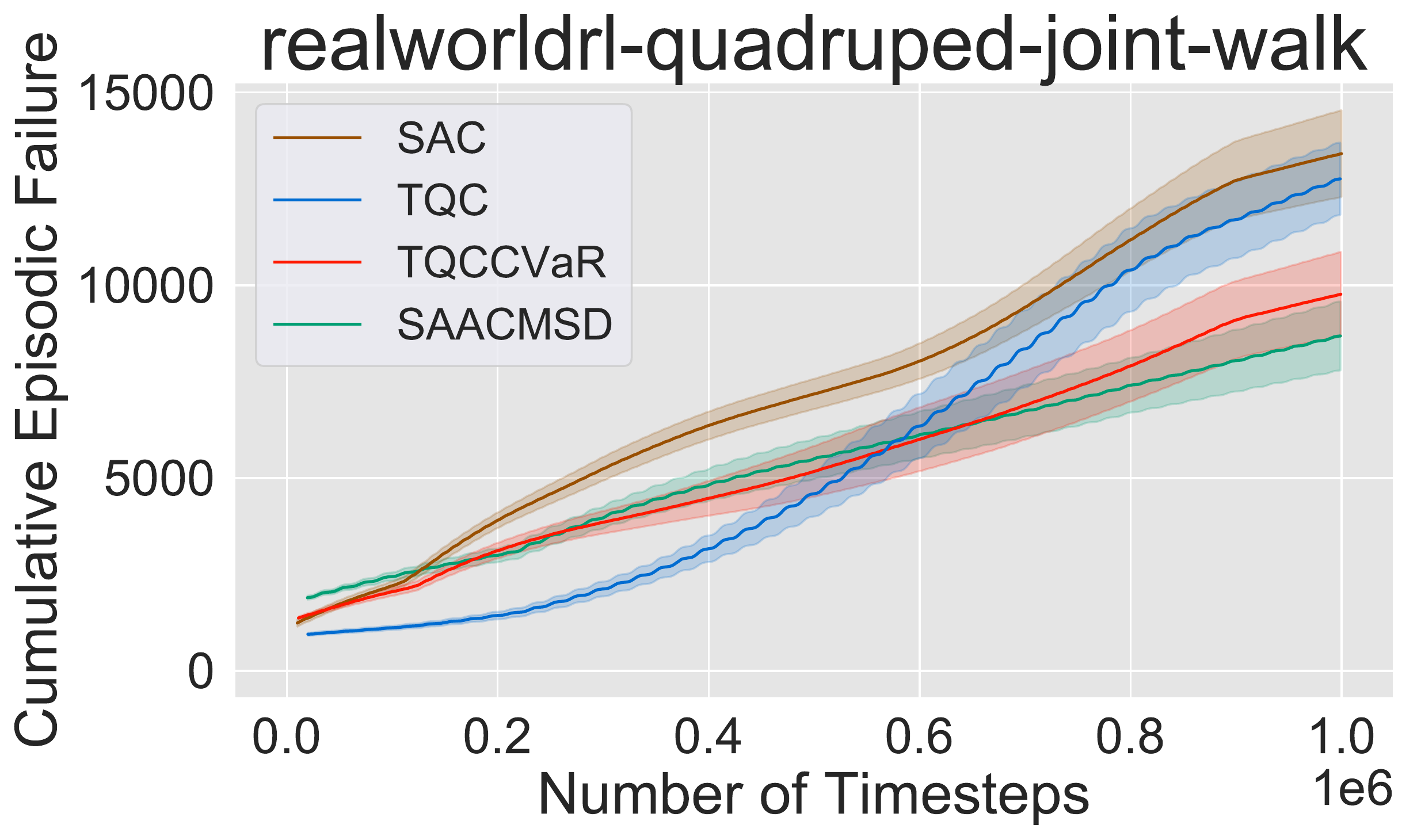}}
    \caption{\algo vs. baselines.}
    \label{fig:constraint3}
\end{minipage}
\end{figure*}

%% file: sections/tables.tex
\begin{table*}[t!]
\centering
\parbox{0.5\linewidth}{
\caption{Comparison of \algo variants.}\label{tab:comparison}
\centering
\resizebox{0.48\textwidth}{!}{
\begin{tabular}{c|c|c}
Method & Efficiency (xSAC) & \# Failures $\pm \sigma$ \\
\hline
SAC & $\times1$ & $65.88 \pm 17.25$ \\
\algocons  & $\times1.33$ & $48.66 \pm 15.99$\\
\algocvar  & $\times2.02$ & $54.39 \pm 15.37$\\
\algomsd  & $\mathbf{\times2.21}$ & $\mathbf{19.31 \pm 3.02}$
\end{tabular}}}\hfill
\parbox{0.5\linewidth}{
\caption{In \textit{quadruped-upright-walk}.}\label{tab:quadruped-upright}
\centering
\resizebox{0.5\textwidth}{!}{
\begin{tabular}{c|c|c}
Method & Efficiency (xSAC) & \# Failures $\pm \sigma$ \\
\hline
SAC & $\times1$ & $8443.93 \pm 696.47$ \\
TQC  & $\times0.97$ & $8297.63 \pm 697.88$\\
TQC-CVaR  & $\times1.03$ & $6298.33 \pm 1078.50$\\
\algomsd  & $\mathbf{\times1.19}$ & $\mathbf{4632.80 \pm 657.35}$
\end{tabular}}}\\
\vspace{-.2em}
\parbox{\linewidth}{
\caption{In \textit{quadruped-joint-walk}.}\label{tab:quadruped-joint}
\centering
\resizebox{0.5\textwidth}{!}{
\begin{tabular}{c|c|c}
Method & Efficiency (xSAC) & \# Failures $\pm \sigma$ \\
\hline
SAC & $\times1$ & $12583.43 \pm 997.29$ \\
TQC  & $\times1.07$ & $11738.57 \pm 995.62$\\
TQC-CVaR  & $\times1.05$ & $9015.82 \pm 1011.31$\\
\algomsd  & $\mathbf{\times1.27}$ & $\mathbf{8069.45 \pm 803.42}$
\end{tabular}}}\vspace*{-1em}
\end{table*}

%% file: sections/conclusion.tex
\section{Discussion and Future Work}
In this paper, we address the problem of risk-sensitive RL under safety constraints and coherent risk measures. We propose that maximizing the value function under risk or safety constraints is equivalent to playing a risk-sensitive non-zero sum (RNS) game. In the RNS game, an adversary tries to maximize the risk of a decision trajectory while the agent tries to maximize a weighted sum of its value function given the adversary's feedback. Specifically, under the MaxEnt RL framework, this RNS game reduces to deploying two soft-actor critics for the agent and the adversary while accounting for a repulsion term between their policies. This allows us to formulate a duelling SAC-based algorithm, called \algo. We instantiate our method for subspace, mean-standard deviation, and CVaR constraints, and also experimentally test it on various continuous control tasks. Our algorithm leads to better risk-sensitive performance than SAC and the risk-sensitive distributional RL baselines in all these environments.
In future work, further study on leveraging the flexibility of \algo to incorporate more safety constraints is anticipated.